\documentclass[11pt]{article}

\PassOptionsToPackage{hyphens}{url}
\PassOptionsToPackage{hidelinks,breaklinks=true}{hyperref}
\usepackage{acl}

\usepackage{times}
\usepackage{latexsym}
\usepackage[T1]{fontenc}
\usepackage[utf8]{inputenc}
\usepackage{microtype}
\usepackage{graphicx}
\usepackage{booktabs}
\usepackage{multirow}
\usepackage{float}      
\usepackage{placeins}   
\usepackage{amsmath}
\usepackage{amssymb}
\usepackage{xcolor}
\usepackage{xspace}

\graphicspath{{figures/}}

\newcommand{\sourceall}{\texttt{source\_all}\xspace}
\newcommand{\effect}{\ensuremath{\Delta}\xspace}

\title{Priors Persist Through Suppression:\\A Stroop Paradigm for Lexical Override}

\author{Han-yu Wang \\
  The University of Hong Kong \\
  \texttt{henry.why@connect.hku.hk}}

\begin{document}
\maketitle

\begin{abstract}
Glossaries, technical specifications, and system prompts routinely ask language models to use
familiar words in unfamiliar ways. The instruction competes with what the word already means, and
even when it wins, the pretrained prior keeps operating underneath. We test this with a
Stroop-style paradigm: a prompt redefines a word (\textit{doctor} now means \textit{forest}),
asks for a related word, and we score the new meaning against the word's pretrained associate
(\textit{hospital}) under matched neutral controls. Across $11$ open-weight models from $1$B to
$9$B parameters, the old meaning interferes in every model, remapping type, and prompt framing we
test. After item-level controls, a model's ordinary preference for the old associate predicts the
size of the interference, in the three document-relevant remapping types though not in antonyms.
On antonym remapping, activation patching in five models locates the repair: restoring three
prompt positions (where the word is redefined, where its new meaning appears, and where the
question repeats it) recovers almost all of the effect (normalized recovery
$R \in [0.92, 1.06]$). The repair is asymmetric. The old meaning's logit falls under any
perturbation of those positions, so pushing it down is not what separates a working override from
a failing one; the new meaning survives only while the position carrying it is intact. What a
local definition achieves is a protected new meaning, not a suppressed old one.
\end{abstract}

\section{Introduction}
\label{sec:intro}

Language models are routinely asked to follow local definitions that temporarily change
what familiar words mean. A legal document may define an everyday term in an unusual way; a
technical manual may repurpose a common word as a domain symbol; a system prompt may introduce
a fictional rule. A software glossary defining \emph{port} as a socket still has to fight the
model's prior preference for \emph{harbor}, and the glossary does not always win. These cases require \emph{contextual override}: the model must use the
locally specified meaning while suppressing interference from the word's pretrained lexical
prior. Override failures are quiet, a residual bias toward the prior in next-token
probabilities rather than a visibly wrong answer, which makes them easy to miss in the settings
where local-definition fidelity matters most, from glossary lookup to retrieval-augmented
document grounding.

Contextual override against lexical priors is \emph{not} a clean replacement of meaning. It
is a competition. This makes two predictions. Items whose distractor is more strongly favored by the model's ordinary
lexical prior should show larger interference, even after item-level controls. And if the rule
wins by installing a local meaning rather than by erasing the old one, then restoring the
positions that carry that meaning should recover the override, while restoring the same word
with someone else's meaning should not. Both predictions are read on one axis, the lexical-prior
\emph{channel} $\log P(t) - \log P(d)$ at the position where the answer is produced.

\begin{figure}[t]
\centering
\includegraphics[width=0.85\columnwidth]{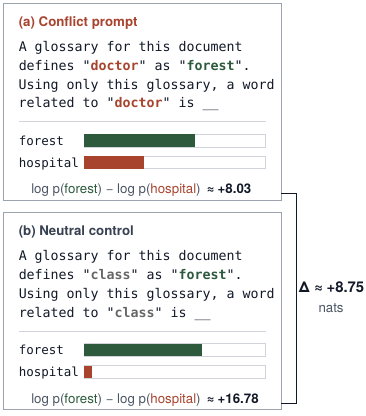}
\caption{The Stroop-style paradigm. Conflict prompt: \emph{doctor means forest} with target
\emph{forest} and lexical-prior distractor \emph{hospital}. Neutral control: query word
replaced by \emph{class} (one of six sampled neutrals). Numbers are sum log-probabilities of
target and distractor under Gemma-2-2B-Instruct (glossary prompt); $\effect$ is neutral minus
conflict $\log P(t) - \log P(d)$ (\S\ref{sec:paradigm}). Displayed neutral gives
$\effect \approx 8.75$; six-control mean $\approx 8.61$.}
\label{fig:paradigm}
\end{figure}

We test these claims with a Stroop-style paradigm (Figure~\ref{fig:paradigm}). A prompt remaps
a query word $w$ to a contextual target $t$; a matched neutral control prompt replaces $w$
with a semantically weak word but holds the target and a competing lexical-prior distractor
$d$ fixed. The target--distractor pair is identical in surface form and tokenization across
conditions, and only the query word changes by design. The structure mirrors the classic
\citet{stroop1935} task, in which a task-relevant dimension (the contextual mapping) is
pitted against an automatic but irrelevant one (the lexical prior). The analogy is
methodological, a matched-control factorial design, and not a claim that a lexical prior is a
competing percept in the cognitive sense.

Behaviorally, across $11$ open-weight models spanning the Qwen, Gemma, OLMo, and Mistral
families and four prompt wrappers, aggregate interference is positive in every model and in
every conflict family. The effect is smallest for the canonical Stroop case (antonym
remapping) and $2$--$4\times$ larger for the arbitrary, polysemy, and domain-definition
families. After item-level controls for answer prior, frequency, tokenization, prompt wrapper,
conflict family, model family, scale, and instruction tuning, the distractor's lexical-prior
strength remains a positive predictor of interference, and the coefficient holds its sign under
clustered and mixed-effects standard errors. The relation is carried by the three
document-relevant families; within antonym remapping alone it is null.

Mechanistically, on five models where activation patching is tractable, we locate where the
override is carried and how the repair reshapes the readout. Patching restores clean activations
into a conflicted run. Three source positions jointly carry the redefinition (the definition
subject, its new target, and the query word); restoring them recovers nearly all of the effect
and exceeds every smaller combination when all are compared at the same site and layer, an
advantage that survives held-out split validation. A definition-target swap, which keeps the redefined word
but replaces its new meaning with another item's, abolishes the recovery, so these positions
bind the local meaning rather than merely matching the repeated word. Decomposing the readout
then shows how the repair works. The distractor loses ground whenever any source position is
disturbed, matched or not, so suppression is a broad side effect. The contextual target is held
up only when the redefinition is restored intact, and that is what binding contributes. This
analysis runs on the antonym family, so it characterizes the repair rather than the predictor.

That language models sometimes fail to override priors is established at the benchmark level, in
few-shot demonstrations, and in entity-level knowledge conflicts \citep{longpre2021entity,
wei2023larger, xu2024knowledge}. What a lexical-level paradigm adds is resolution. Because the
target and distractor are held fixed while only the query word moves, interference becomes a
quantity attached to a single item, and that quantity, $\log P(t) - \log P(d)$, serves as both
the behavioral metric and the patching score. Measuring behavior and mechanism on one channel is
a design decision rather than a result: it buys commensurability between the two, while the
evidence on each side still has to stand on its own. Concretely, we contribute
(i) a controlled Stroop-style paradigm for lexical-level override, with matched neutral baselines
that isolate lexical-prior interference per item; (ii) behavioral evidence across $11$
open-weight models that lexical-prior strength predicts override failure after item-level
controls; and (iii) a mechanistic
localization in which a source-position triplet recovers the override while the binding-specific
effect, target preservation, narrows to the definition-target position.

\section{Related Work}
\label{sec:related}

\paragraph{Conflicts between context and parametric knowledge.}
A growing body of work studies how language models reconcile in-context information with
information memorized at pretraining \citep{longpre2021entity, xie2024knowledge,
kortukov2024studying}, with \citet{xu2024knowledge} surveying context--memory, inter-context,
and intra-memory conflicts. Almost all of it operates at the document or claim level (entity
reassignment, passage contradiction) and scores open generation. We move the conflict down to
the lexical level, where a matched neutral remapping holds the target/distractor pair fixed and
makes context--memory conflict measurable per item and testable per position.

\paragraph{Psycholinguistic probes of lexical interference.}
Earlier probes measure lexical interference without Stroop framing:
\citet{ettinger2020bert} adapts cloze diagnostics to pretrained models;
\citet{kassner2020negated} show that a misleading prime drags masked-token predictions toward
the prime; \citet{misra2020priming} find that semantic priming in BERT flips to interference
under stronger sentential constraint. We induce the conflict with an explicit local definition
rather than a freestanding prime; this turns lexical-prior advantage into a quantitative
item-level predictor of interference (\S\ref{sec:behavior}), and the matched neutral controls
supply the baseline that a freestanding prime leaves implicit.

\paragraph{In-context learning and prior tug-of-war.}
\citet{wei2023larger} and \citet{pan2023incontext} document a tension between
\emph{task recognition} from priors and \emph{task learning} from demonstrations, with scale
and instruction tuning shifting the balance; \citet{shi2023distracted} show that irrelevant
context pulls predictions away from the prior. Cast as implicit Bayesian inference
\citep{xie2022explanation}, this is a competition between a prior over the query word and an
in-context likelihood supplied by the rule. A single in-prompt redefinition replaces multi-shot
demonstration and a binary lexical readout replaces task accuracy, so prior strength is measured
directly per item with the target/distractor pair held fixed.

\paragraph{Mechanistic interpretability of knowledge conflicts.}
Activation patching \citep{vig2020causal, meng2022rome, zhang2024towards} (an instance of
causal mediation analysis, \citealp{mueller2024quest}) and circuit-level analyses
\citep{wang2023ioi, olsson2022induction, mcdougall2024copy} localize internal representations
causally, with copy suppression \citep{mcdougall2024copy} establishing that
attention heads can shape outputs by lowering wrong-token logits rather than by writing
correct ones. Within knowledge conflicts, \citet{yu2023characterizing},
\citet{ortu2024competition}, and \citet{jin2024cutting} identify attention heads that promote
memorized versus in-context answers at the document and entity level, taking the individual
head as the unit of analysis. We instead take a source-position triplet (definition subject,
definition target, query word) as the unit and ask how a single in-prompt redefinition competes
with a word's pretrained prior.

\paragraph{Binding and representation engineering.}
Existing binding work treats in-context use as keeping entities paired with the right
attributes: \citet{feng2024binding} identify binding-ID vectors attached to entities and
attributes, while \citet{prakash2024finetuning} show that fine-tuning improves entity tracking
largely by strengthening mechanisms already present in the base model. Lexical override adds a
source of competition that entity binding does not face: the model is not retrieving an attribute
supplied in context but reusing a familiar word whose pretrained meaning remains a plausible
answer.
Representation-engineering methods
\citep{turner2023activation, zou2023representation, hernandez2024inspecting} steer or edit
hidden states through model-side directions or fact encodings; our intervention instead restores
a model's own clean activations at chosen source positions, locating which position carries
the new meaning rather than imposing a direction.

\section{The Stroop-style Paradigm}
\label{sec:paradigm}

\subsection{Stimuli}

Each item specifies a query word $w$, a lexical-prior distractor $d$, and a context-defined
target $t$. The conflict prompt remaps $w$ to $t$ inside one of four prompt wrappers (defined
below); the neutral prompt replaces $w$ with a semantically weak word $n$ chosen from a
neutral pool, while leaving $t$ and $d$ unchanged (Figure~\ref{fig:paradigm}). For the
running \emph{doctor} item under the glossary wrapper, the conflict prompt reads
\texttt{A glossary for this document defines "doctor" as "forest". Using only this glossary,
a word related to "doctor" is}, with $t={}$\emph{forest}, $d={}$\emph{hospital}; the neutral
prompt substitutes a neutral $n$ for \emph{doctor}. The relation cue is family-specific (\texttt{a synonym of} for antonym,
\texttt{a word related to} for the other three; Appendix~\ref{app:stimuli},
Table~\ref{tab:prompt_templates}). Target and distractor completions are identical in surface
form and tokenization across conditions; the query word varies by design. We average over
multiple neutral words per item, and the controlled regression includes token-count and frequency
covariates (\S\ref{sec:lexical_predictor}).

\paragraph{Conflict families.}
We evaluate four conflict families that span psycholinguistic and document-style settings:
\textbf{(i)} \emph{antonym remapping} (\emph{small means big}; the original Stroop analogue);
\textbf{(ii)} \emph{arbitrary semantic remapping} (\emph{apple means river}); \textbf{(iii)}
\emph{polysemy/entity remapping} (\emph{jaguar means car}); and \textbf{(iv)}
\emph{domain-definition remapping} (\emph{thread means process}). Per-family examples in
Table~\ref{tab:stim_examples}.

\paragraph{Prompt wrappers.}
Each item is rendered under four wrappers: game-rule (toy framing), glossary (document
glossary), technical-document (technical convention), and scoped-definition (passage-scoped
rule); full templates in Table~\ref{tab:prompt_templates}. This tests whether interference is
tied to toy wording or persists in document-like settings.

\subsection{Behavioral metric}

For prompt $x$, let $S(x) = \log P(t \mid x) - \log P(d \mid x)$ denote relative target preference,
where $P(\cdot \mid x)$ is the model's full-sequence probability under teacher forcing.
The Stroop-style interference effect is
\[
\effect = \mathbb{E}_{n \in \mathcal{N}}\!\left[ S(x_{\text{neutral},n}) \right] - S(x_{\text{conflict}}),
\]
where $\mathcal{N}$ is the set of sampled neutral controls for the item.
Positive \effect~means lexical conflict reduces relative target preference.
We report sum log-probability as the primary metric. Mean per-token log-probability gives a
length-normalized diagnostic and yields the same conclusions at the model level, where all $11$
settings stay positive ($1.32$ to $2.65$), and in the controlled regression. One of the $44$
model$\times$family cells changes sign between the two scoring modes (Mistral-7B-IT on antonym
remapping, $+0.34$ under sum and $-0.59$ under mean), and none of the $44$ model$\times$wrapper
cells does.

\paragraph{Worked example.}
For the running \emph{doctor} item (Figure~\ref{fig:paradigm}; Gemma-2-2B-Instruct under the
glossary wrapper, sum log-probability), $S(x_{\text{conflict}}) \approx {+}8.03$. The displayed
neutral control using \emph{class} gives $S(x_{\text{neutral,\,class}}) \approx {+}16.78$ and
a single-control $\effect \approx 8.75$; the item-level metric averages over six neutral
controls, giving mean $S(x_{\text{neutral}}) \approx {+}16.64$ and $\effect \approx 8.61$.

\paragraph{Lexical-prior strength.}
We measure the strength of the distractor's lexical prior with an
\emph{ordinary-prior prompt} (e.g., \emph{a word related to doctor is}) that contains no
remapping rule. The relation cue matches the family of the conflict prompt, so the prior is
measured under the same relation phrase the conflict prompt uses. The lexical-prior advantage
of the distractor over the target is $\log P(d) - \log P(t)$ under that prompt, computed per
item per model from a single forward pass. This quantity is what we mean by lexical-prior
strength throughout.

\subsection{Models}

The behavioral analysis evaluates $11$ open-weight model settings: Qwen2.5-1.5B and 7B base/instruct
\citep{yang2024qwen25}, Gemma-2-2B and 9B base/instruct \citep{riviere2024gemma2},
OLMo-1B \citep{groeneveld2024olmo}, and Mistral-7B base/instruct \citep{jiang2023mistral}.
We abbreviate the instruction-tuned variants as ``-IT'' in figures and tables.
Mechanistic analysis uses the tractable subset for which full activation patching with
TransformerLens \citep{nanda2022transformerlens} is computationally feasible:
Qwen2.5-1.5B base/IT, Gemma-2-2B base/IT, and OLMo-1B (Table~\ref{tab:model_set}).

\section{Behavioral Results}
\label{sec:behavior}

\subsection{Interference is positive across modern models}

\begin{figure}[t]
\centering
\includegraphics[width=\columnwidth]{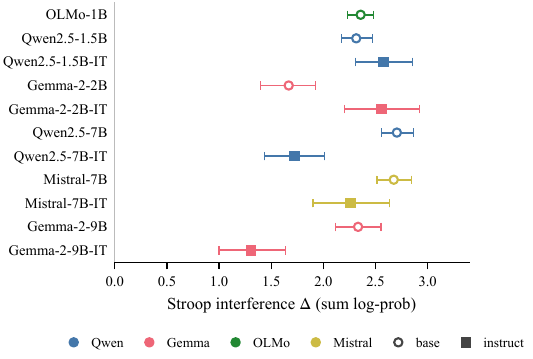}
\caption{Model-level Stroop interference \effect~with bootstrap $95\%$ confidence intervals
(sum log-probability scoring). All $11$ evaluated model settings are positive at the aggregate
level, including $7$B/$9$B-class models from four model families.}
\label{fig:forest}
\end{figure}

\begin{figure*}[t]
\centering
\includegraphics[width=\textwidth]{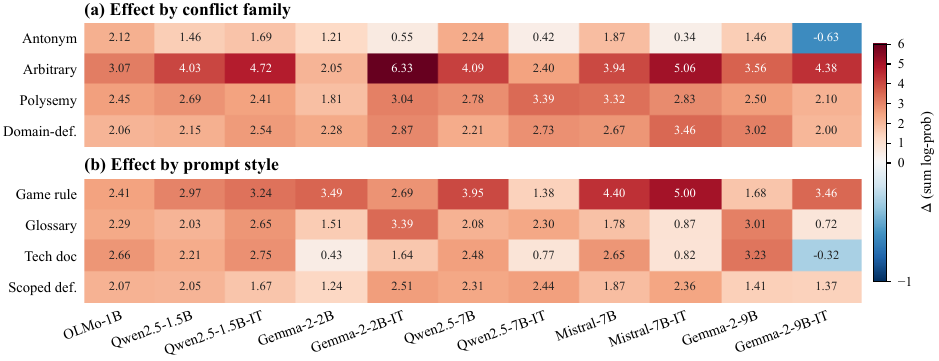}
\caption{External validity. \textbf{(a)} Effect by conflict family across models.
\textbf{(b)} Effect by prompt wrapper across models. The effect is positive in every family, and
the document-relevant families (arbitrary, polysemy, domain-definition) produce larger effects
than antonym remapping, which is also the only family with a
negative cell. The effect is positive in aggregate under all document-style wrappers. Cell
values are sum log-probability \effect.}
\label{fig:heatmaps}
\end{figure*}

All $11$ settings have positive aggregate effects (Figure~\ref{fig:forest}), with effect sizes
ranging from $1.31$ (Gemma-2-9B-IT) to $2.71$ (Qwen2.5-7B). Interference is not specific to
small models or to a single training regime, and it appears in all four model families.

\subsection{The effect generalizes across conflict families}

The effect is positive in every conflict family. Antonym remapping, the closest psycholinguistic
analogue of a Stroop conflict, gives the smallest aggregate effect
($\effect = 1.16$, $95\%$~CI $[1.06, 1.26]$, $n{=}3520$; Figure~\ref{fig:heatmaps}(a)), and the
three document-relevant families are $2$--$4\times$ larger: arbitrary semantic remapping
$3.97$\,$[3.79, 4.15]$, polysemy/entity $2.66$\,$[2.49, 2.84]$, domain-definition
$2.54$\,$[2.37, 2.73]$. The antonym minimum sits in instruction-tuned models. The family's four smallest
cells are Gemma-2-9B-IT ($-0.63$, the only negative cell in the family), Mistral-7B-IT ($0.34$),
Qwen2.5-7B-IT ($0.42$), and Gemma-2-2B-IT ($0.55$), against $1.21$ to $2.24$ across the seven
remaining settings; the pattern is not size-monotone, since a $2$B instruction-tuned model falls
in the low group while a $1.5$B one ($1.69$) does not (Table~\ref{tab:model_x_family}). Antonym
is also the only family whose relation cue is synonym-seeking, a confound taken up in
Limitations. The three document-relevant families are the settings where lexical override
matters in deployment: glossary lookup, polysemous-entity disambiguation, and domain-specific
redefinition.

\subsection{The effect persists under document-style wrappers}

Interference survives reframing the conflict as a glossary entry, a technical convention, or a
passage-scoped definition. The game-rule wrapper yields the largest aggregate effect ($3.15$),
followed by glossary ($2.06$), scoped-definition ($1.94$), and technical-document ($1.75$), a
$1.8\times$ range in mean magnitude (Figure~\ref{fig:heatmaps}(b)). That ordering is not matched
by the rate at which interference appears: the fraction of positive item--prompt observations is
$78$--$81\%$ under all four wrappers. Document-style framing changes how much an override costs
rather than how often it is paid. Of $44$ model$\times$wrapper cells, $42$ have CI $> 0$, and the
single negative point estimate is Gemma-2-9B-IT under the technical-document wrapper ($-0.32$;
Table~\ref{tab:model_x_style}).

\subsection{Lexical-prior strength predicts interference}
\label{sec:lexical_predictor}

\begin{figure}[t]
\centering
\includegraphics[width=\columnwidth]{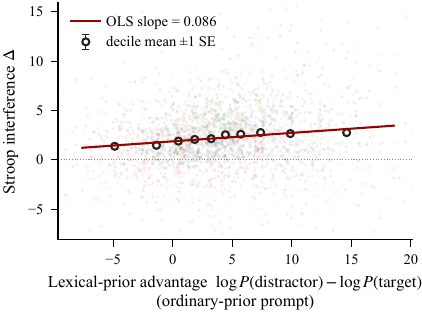}
\caption{Lexical-prior strength predicts Stroop interference at the item$\times$prompt level.
Light points: $2{,}500$ subsampled item$\times$prompt observations across all $11$ models.
Open circles: decile means with $\pm 1$\,SE bars. Red line: bivariate OLS fit
($n{=}7{,}744$). The controlled regression in Table~\ref{tab:regression} retains a positive
coefficient after adding answer prior, tokenization, frequency, prompt wrapper, conflict family,
model family, scale, and instruction tuning.}
\label{fig:lexical}
\end{figure}

If interference reflects lexical-prior conflict rather than incidental confounds, then items
whose distractor is more strongly favored by the model's ordinary lexical prior should show
larger \effect. They do (Figure~\ref{fig:lexical}). Pooling all $11$ models at the
item--prompt level, bivariate OLS gives a slope of $0.086$ ($p < 10^{-9}$), and binned decile
means trace a monotone trend: one additional nat of lexical-prior advantage raises
interference by $0.086$ nat. Lexical-prior advantage ranges over more than $20$ nats across
items (Figure~\ref{fig:lexical}), so moving from one end of that range to the other shifts
predicted interference by roughly $1.7$ nats, a swing of the same order as the model-level
aggregate effects themselves ($1.31$ to $2.71$; Table~\ref{tab:per_model}).

\begin{table}[t]
\centering
\scriptsize
\setlength{\tabcolsep}{3pt}
\begin{tabular*}{\columnwidth}{@{\extracolsep{\fill}}lrrr@{}}
\toprule
Term & Coef. & $95\%$ CI & $p$ \\
\midrule
Intercept                 & $+0.746$ & $[+0.38, +1.11]$ & $5.5{\cdot}10^{-5}$ \\
Lexical advantage         & $+0.114$ & $[+0.08, +0.15]$ & $4.5{\cdot}10^{-10}$ \\
Answer prior              & $+0.16$ & $[+0.11, +0.21]$ & $7.7{\cdot}10^{-11}$ \\
Frequency difference (zipf) & $+0.65$ & $[+0.43, +0.86]$ & $7.9{\cdot}10^{-9}$ \\
Token-count difference    & $-0.08$ & $[-0.46, +0.31]$ & $0.69$ \\
Is instruct               & $-0.74$ & $[-1.00, -0.47]$ & $4.0{\cdot}10^{-8}$ \\
$\log$ params (B)         & $-0.32$ & $[-0.44, -0.19]$ & $4.3{\cdot}10^{-7}$ \\
\bottomrule
\end{tabular*}
\caption{Main controlled regression of item$\times$prompt Stroop interference under sum
log-probability scoring ($n{=}7{,}744$). The model also includes fixed effects for conflict
family, prompt wrapper, and model family (omitted for space; see Appendix~\ref{app:regression}).
Standard errors are clustered by item; Table~\ref{tab:regression_robustness} reports three
alternative specifications.}
\label{tab:regression}
\end{table}

Table~\ref{tab:regression} reports the controlled OLS regression.
After accounting for answer prior (whether the target outscores the distractor with no
remapping), frequency, tokenization, prompt wrapper, conflict family,
model family, scale, and instruction tuning, the lexical-advantage coefficient remains positive
($+0.114$, $p < 10^{-9}$) and the intercept remains positive ($+0.746$, $p < 10^{-4}$). The
positive intercept matters for how the predictor should be read: interference does not vanish at
zero lexical-prior advantage, so prior strength modulates a baseline effect rather than
accounting for all of it.
The instruct and $\log$-parameter coefficients are negative on average, and the average masks a
scale interaction. Pairwise base-versus-IT differences in the model-level aggregate effect show
instruction tuning \emph{raising} interference at $\leq 2$B ($+0.26$ to $+0.89$) and
\emph{lowering} it at $\geq 7$B ($-0.41$ to $-1.02$), computed from the unrounded model-level
means that Table~\ref{tab:per_model} reports to two decimals.

Standard errors in the main table are clustered by item. The point estimate is identical under
clustering by model, two-way clustering by item and model, and a linear mixed-effects
specification with item- and model-level random effects, but the interval widens
across them: the $95\%$ CI runs from $[+0.08, +0.15]$ under item-clustering to
$[+0.01, +0.21]$ under mixed effects, where $p = 0.025$ (Table~\ref{tab:regression_robustness}).
The sign of the coefficient is stable across all four specifications; its magnitude is well
determined only under item-clustering, since the mixed-effects lower bound ($+0.014$) admits an
effect nearly an order of magnitude smaller than the point estimate. The instruction-tuning and
scale coefficients likewise keep their negative sign with wider intervals under model-level and
mixed-effects specifications.

The pooled coefficient is not uniform across conflict families. In the
$\textsc{lexical\_advantage} \times \textsc{conflict\_family}$ specification the antonym slope is
null ($+0.026$, $p = 0.235$), while arbitrary semantic, domain-definition, and polysemy/entity
remapping give $+0.202$, $+0.185$, and $+0.150$ (all $p < 10^{-3}$;
Table~\ref{tab:regression_interactions}). The pooled effect is carried by the three
document-relevant families, which are also the three that use the broader relation cue
(Limitations). What the controls establish is correspondingly narrow. The answer-prior covariate
holds context-free target ease fixed and the fixed effects hold prompt wrapper, conflict family,
and model family constant, so the surviving coefficient is a within-cell association rather than
a family- or template-level artifact. The token-count covariate is itself null ($-0.08$,
$p = 0.69$): residual differences in target and distractor tokenization do not track interference.

\section{Mechanistic Analysis}
\label{sec:mechanism}

The behavioral effect is broad, and it survives the surface controls we can measure. What those
controls cannot show is how the conflict is resolved inside the network: where the local definition is carried, and how restoring it reshapes the
target-versus-distractor readout.

\paragraph{Scope.}
Mechanistic experiments use the antonym-remapping family, the game-rule prompt wrapper, and the
single-token subset, on five models for which TransformerLens loading is reliable. Antonym is the
family whose per-item prior slope is null (\S\ref{sec:lexical_predictor}), so this section asks
how an override is repaired rather than why prior strength predicts which overrides fail. The two
questions share a quantity, the readout margin $\log P(t) - \log P(d)$ that defines both the
behavioral metric and the patching score, but joining them into a single account assumes a
generalization across conflict families that we do not test (Limitations).

\subsection{Setup: patching, conditions, and recovery}
\label{sec:mech_setup}

For each item we construct a clean prompt (neutral remapping) and a corrupted prompt
(lexical-conflict remapping) sharing the same target and distractor. An item is eligible when its
query word, target, and distractor are single tokens, when the two prompts match in token length,
and when the item shows positive interference for that model, which is what leaves recovery a
gap to close. This gives $32$ items per model and $14$ for Gemma-2-2B-IT
(Appendix~\ref{app:mech_impl}). Activation patching copies
clean residual-stream activations into the corrupted run at chosen positions and measures how
far this restores the clean answer. The three source positions we patch are the
\emph{definition subject} (e.g.\ \emph{small}), the \emph{definition target} (e.g.\ \emph{big}),
and the \emph{query word}, taken individually, in pairs, and as the \emph{full source triplet}
of all three. The clean and corrupted prompts differ at exactly two token positions, the
definition subject and the query word; the definition-target position holds the same token in
both runs, so patching it transfers only what that position has accumulated from context rather
than a difference in token identity. The score is the standard normalized recovery $R$ of
\citet{zhang2024towards} on
the final-position target-minus-distractor logit difference: $R{=}1$ is full clean recovery,
$R{=}0$ is none, and $R$ can leave $[0,1]$ when a patch overshoots or actively harms the answer.
The clean$-$corrupt gap in this margin is the mechanistic counterpart of the behavioral
interference \effect, so $R{=}1$ means the patch fully closes it.

We read the triplet as filling three slots: the subject and query word together fix the
\emph{identity match} (which word is being redefined), while the definition target supplies the
\emph{local value} (what it now means). We call the triplet's job \emph{binding}: making the
readout use this local value in place of the word's pretrained meaning. Two corruption
conditions test this reading against the
alternative that recovery merely reflects repeated words. A \emph{definition-target swap} keeps
the subject and query word matched to the item but replaces only the definition-target
activation with a donor item's, preserving the identity match while corrupting the local value.
An \emph{item-mismatched control} instead patches all source activations from a randomly chosen
different item. Appendix~\ref{app:mech_impl} gives the implementation details.

\subsection{The source triplet recovers the override effect}
\label{sec:mech_recovery}

\begin{figure}[t]
\centering
\includegraphics[width=\columnwidth]{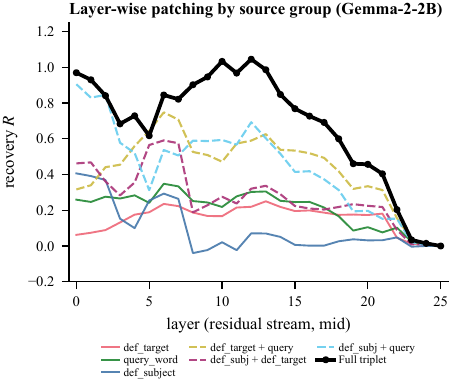}
\caption{Layer-wise patching recovery for each source group on Gemma-2-2B, at the
residual-stream site the full triplet selects (mid-block). The full triplet sits above every
singleton and every pair through mid-network depths. Recoveries for all five mechanism models,
each at its own selected site and layer, are in Table~\ref{tab:mechsummary}.}
\label{fig:mechcore}
\end{figure}

\begin{table}[t]
\centering
\scriptsize
\setlength{\tabcolsep}{3pt}
\begin{tabular*}{\columnwidth}{@{\extracolsep{\fill}}lllrrr@{}}
\toprule
Model & Site & $L$ & $R_{\text{trp}}$ & $R_{\text{pair}}$ & $\Delta R$ \\
\midrule
Qwen2.5-1.5B    & input     & 3  & $0.99$ & $0.89$ & $+0.11$ \\
Qwen2.5-1.5B-IT & input     & 15 & $0.92$ & $0.59$ & $+0.33$ \\
Gemma-2-2B      & mid-block & 12 & $1.05$ & $0.69$ & $+0.35$ \\
Gemma-2-2B-IT   & input     & 3  & $1.04$ & $0.68$ & $+0.36$ \\
OLMo-1B         & input     & 2  & $1.06$ & $0.87$ & $+0.19$ \\
\bottomrule
\end{tabular*}
\caption{Source-triplet residual patching versus the best source-position pair. Site
(residual-stream location: input = block input, mid-block = post-attention) and $L$ (layer)
are selected by the full source triplet; $R_{\text{trp}}$ is the recovery of the full triplet
at that site/layer, and $R_{\text{pair}}$ is the recovery of the best source-position pair at
the same site/layer. The best pair is consistently definition subject + query word; adding
the definition target improves point-estimate recovery in all five models.}
\label{tab:mechsummary}
\end{table}

Joint patching of the definition subject, definition target, and query word recovers nearly
all of the override effect (aggregate $R \in [0.92, 1.06]$ across five models, mean $1.01$),
exceeding any source-position pair at the residual-stream site and layer where the triplet's
recovery peaks (the position it \emph{selects}; Figure~\ref{fig:mechcore},
Table~\ref{tab:mechsummary}; triplet--pair margin
$\Delta R = R_{\text{triplet}} - R_{\text{pair}} \in [+0.11, +0.36]$ in 5/5 models). The
best pair is consistently definition subject + query word; adding the definition target
improves recovery at the same site and layer. Recovery for every source
group at this site/layer is in Appendix~\ref{app:mech_extra},
Table~\ref{tab:source_group_recovery}.

The comparison is made at one site and layer per model, the point at which the triplet's own
recovery peaks, and that choice matters. Evaluated instead at its own best site and layer, the
definition subject + query word pair reaches high recovery in early layers, where the patch
behaves like an input replacement rather than like a targeted restoration
(Appendix~\ref{app:mech_extra}). The triplet advantage is therefore a statement about a common
comparison point rather than about every protocol. The split validation below is the test that
does not inherit this choice, because it selects site, layer, and source group on discovery items
and scores the margin on items held out from that selection.

The recovery is item-specific. Across all $250$ control trials ($50$ item-mismatched
perturbations per model $\times$ $5$ models), clean source activations drawn from a different
item yield strongly \emph{negative} recovery ($-0.58$ to $-4.16$;
Table~\ref{tab:random_summary}); no single trial reaches positive recovery and none approach
the matched value, so recovery requires source information matched to the item rather than
generic clean activations.

Across $200$ discovery/held-out splits, the held-out triplet--pair margin has a positive mean in
all five models ($+0.11$ to $+0.39$) and is positive in $0.97$--$1.00$ of splits for the three
base models. The instruction-tuned models are more heterogeneous ($0.95$ and $0.83$ of splits),
and three of the five split-level $95\%$ intervals reach or cross zero at the lower end
(Appendix~\ref{app:mech_extra}, Table~\ref{tab:split_summary}). The margin is reliably signed in
aggregate rather than in every split, least reliably in Gemma-2-2B-IT.

\subsection{Binding localizes to the definition-target position}
\label{sec:mech_swap}

The triplet might recover the effect only because it repeats the redefined word, without
carrying its new meaning. The definition-target swap rules this out: keeping subject and query
word matched while replacing only the definition-target activation collapses recovery in every
item of every model ($142/142$ items; the matched$-$swap gap is at least $1.9$ nats in the
final-position logit margin for every item, and $+1.54$ to $+5.26$ in recovery units at the model
level; Appendix~\ref{app:mech_extra}, Table~\ref{tab:swap_summary}). The local value, not the
repeated identity, is what the triplet must carry. The best pair is the triplet without its
local-value slot, so $\Delta R$ is the cost of leaving that slot unfilled and the swap is the
cost of filling it wrongly.

\begin{table*}[!t]
\centering
\scriptsize
\setlength{\tabcolsep}{3pt}
\begin{tabular*}{\textwidth}{@{\extracolsep{\fill}}lrrrrrrrrr@{}}
\toprule
& \multicolumn{3}{c}{Matched (M)} & \multicolumn{3}{c}{Swap (S)} & \multicolumn{3}{c}{Mismatch (R)} \\
\cmidrule(lr){2-4} \cmidrule(lr){5-7} \cmidrule(lr){8-10}
Model & $\Delta\ell_t$ & $\Delta\ell_d$ & $\Delta m$ & $\Delta\ell_t$ & $\Delta\ell_d$ & $\Delta m$ & $\Delta\ell_t$ & $\Delta\ell_d$ & $\Delta m$ \\
\midrule
Qwen2.5-1.5B    & $-1.39$ & $-3.13$  & $+1.74$ & $-12.79$ & $-5.34$  & $-7.45$ & $-12.68$ & $-5.41$  & $-7.27$ \\
Qwen2.5-1.5B-IT & $-6.97$ & $-11.05$ & $+4.08$ & $-16.48$ & $-13.70$ & $-2.77$ & $-16.26$ & $-13.34$ & $-2.91$ \\
Gemma-2-2B      & $+0.17$ & $-3.72$  & $+3.89$ & $-8.04$  & $-6.02$  & $-2.03$ & $-7.95$  & $-5.81$  & $-2.14$ \\
Gemma-2-2B-IT   & $-1.78$ & $-5.60$  & $+3.83$ & $-16.92$ & $-8.03$  & $-8.89$ & $-15.69$ & $-7.40$  & $-8.30$ \\
OLMo-1B         & $-1.38$ & $-3.41$  & $+2.03$ & $-7.68$  & $-5.50$  & $-2.19$ & $-7.56$  & $-5.33$  & $-2.23$ \\
\bottomrule
\end{tabular*}
\caption{Logit decomposition under three patching conditions. M = matched triplet,
S = definition-target swap, R = item-mismatched control. $\Delta m = \Delta\ell_t -
\Delta\ell_d$. The distractor drops under all three conditions (S and R within $0.7$ nat of each
other); the target's drop grows sharply only when binding is disrupted. Margin gaps
$|\Delta m_S - \Delta m_R| \leq 0.6$ nat across all five models.}
\label{tab:logit_decomp_all}
\end{table*}

\subsection{Binding preserves the target rather than suppressing the distractor}
\label{sec:mech_signature}

Why does corrupting the definition target cost so much recovery? Decomposing the margin movement
$\Delta m = \Delta\ell_t - \Delta\ell_d$ splits it into two terms, what happens to the contextual
target's logit and what happens to the prior's distractor's
(Table~\ref{tab:logit_decomp_all}).

Under the matched patch the target logit moves little in four models ($\Delta\ell_t$ between
$-1.78$ and $+0.17$) and falls by $6.97$ nats in Qwen2.5-1.5B-IT, while the distractor falls by
$3.1$ to $11.1$ nats. Under the swap and item-mismatched controls the distractor falls comparably
($5.3$ to $13.7$ nats) and the target falls by $7.6$ to $16.9$ nats, inverting the margin. Read
within each model, corrupting the local value costs the target a further $6.3$ to $15.1$ nats
relative to the matched patch while changing the distractor's loss by only $2.1$ to $2.7$ nats.
The two corruptions are nearly indistinguishable at the margin level
($|\Delta m_S - \Delta m_R| \leq 0.6$ nat across all five models). Distractor suppression is a
broad consequence of perturbing any source position; holding up the contextual target is what
binding contributes, and it localizes to the definition-target position. Component-level scans of
the attention heads and MLPs that read this margin out are in Appendix~\ref{app:mech_extra}.

\section{Discussion}
\label{sec:discussion}

The two analyses answer different questions and meet at one quantity. Behaviorally, a
one-forward-pass estimate of distractor-versus-target prior strength predicts per-item override
interference across $11$ open-weight models, after item-level controls and across four prompt
wrappers including document-style framings, with the association carried by the three
document-relevant conflict families. Mechanistically, on the antonym family, a source-position
triplet recovers the override effect, and that recovery is read on the same
target-minus-distractor margin that defines the behavioral metric. Sharing the margin is what
makes the two results commensurable. It does not by itself show that the repair mechanism
transports to the families where the behavioral predictor is strongest, and that transport is the
main question this design leaves open.

The logit decomposition turns that recovery into a division of labor. The distractor loses ground
under every corruption we ran, structured (the definition-target swap) and unstructured (the
item-mismatched control) alike, so its suppression cannot be what an intact definition
contributes. The contribution has to be the contextual target's survival, which holds only while
the definition-target position is undisturbed. That inverts the intuitive picture. When a
glossary or a system prompt loses to a familiar word, the diagnostic question is not whether the
model suppressed the old meaning hard enough, but whether the redefinition's own representation
survived far enough into the network to be read out.

Partial competition of this kind has a normative counterpart. In sequential belief updating under
a limited compute budget, the resource-rational response is to discount new evidence while
carrying the prior at full weight: the optimal rule raises the likelihood to an exponent below
one but leaves the prior's exponent at one \citep{zhu2025computation}, the asymmetry long
documented behaviorally as conservative updating \citep{edwards1968conservatism,
grether1992testing}. Contextual override asks for the opposite operation, since a local
definition instructs the model to drive the prior's weight toward zero rather than to combine it
with evidence. Our measurements do not estimate exponents, so the correspondence is qualitative:
what they show is the direction such an account predicts, an in-context definition incorporated
only partially while the lexical prior persists at strength. Read this way, override failure is
less a generic robustness defect than the visible cost of a prior-preserving update meeting a
task that demands prior replacement.

\section{Conclusion}
\label{sec:conclusion}

A local definition does not replace what a word already means. It competes with it. Redefining a
familiar word costs the model preference for the meaning that definition assigns, and the cost is
positive in every one of the $11$ open-weight models, four conflict families, and four prompt
wrappers we test. In the three document-relevant families its size tracks how strongly the model
ordinarily prefers the word's lexical-prior distractor.

Inside the network the redefinition is carried by three source positions in the prompt, and
restoring all three recovers the override most completely. The old meaning's logit falls under
any perturbation of these positions, matched or not, so pushing it down is not what distinguishes
a successful override from a failed one. The new meaning survives only while the position that
carries it is intact. Lexical override, on this evidence, is less an act of erasure than an act
of maintenance.

The practical consequence is that the items at risk can be identified in advance. A glossary
entry, a scoped definition in a retrieved passage, or a system prompt that repurposes a common
term is most vulnerable where the word's ordinary associate is strongest, and that strength costs
one forward pass to measure. The failure is quiet by construction here, a shifted preference
between two fixed continuations; what it looks like in free generation is beyond what this design
can say.

The reach of the mechanism is what remains open. The behavioral effect spans $1$B to $9$B
parameters and all four conflict families, while the causal account rests on five $1$B--$2$B
models and on antonym remapping, where the item-level predictor does not hold. Showing that the
same three positions carry document-relevant remappings at larger scale would join the two halves
of this paper into a single account. The paradigm, the stimuli, and the patching code are
released so that the test can be run directly.

\section*{Limitations}

The paradigm is deliberately controlled. Even with glossary, technical-document, and
scoped-definition wrappers, our prompts are shorter than realistic long-form documents, and
the items use English stimuli with target-versus-distractor likelihood scoring rather than
unconstrained free generation.

The mechanistic analysis covers five $1$B--$2$B models where full residual and component patching
is tractable, while the behavioral analysis spans $1$B--$9$B across four conflict families. Both
the source-triplet structure and the target-preservation versus distractor-suppression division
of labor are established on a narrow slice of the stimulus set: antonym remapping, the game-rule
wrapper, the single-token subset, and, within that, only items showing positive interference for
the model in question. The last restriction is forced by the metric, since recovery normalizes by
a clean$-$corrupt gap that an item without interference does not provide. Antonym is also the
family carrying the relation-cue confound discussed below, and the one where the item-level
predictor is null, so the two halves of the paper rest on different parts of the stimulus set.
Whether the same three positions carry the other conflict families, non-game-rule wrappers, and
$7$B$+$ scale is open.

The triplet-versus-pair comparison is additionally made at one site and layer per model
(\S\ref{sec:mech_recovery}); held-out split validation supports the margin without inheriting
that choice, with wider split-level intervals in the instruction-tuned models. The
definition-target swap supports binding only at the
source-representation level; a same-target donor swap (donors constrained to share the target
word, isolating context-binding from surface-form effects) and a complete path-level circuit
through writer/reader paths remain future work.

The antonym family is structurally constrained: a synonym-seeking cue (\texttt{a synonym of})
is the only phrasing that keeps distractor and remap target separable for antonym pairs,
since broader cues would invite the antonym itself as a high-prior continuation and collapse
the conflict. This cue may activate an opposite-word routine in instruction-tuned models, which
would fit the antonym minimum observed in the instruction-tuned settings, but we do not test it
directly. The other three families use the broader relation cue and are not exposed to this
particular confound. Disentangling the routine would require varying the synonym-cueing wording
while preserving conflict structure, which we leave to future work.

\section*{Ethical Considerations}

This work uses synthetic stimuli and open-weight models; no human subjects or sensitive data are
involved. The paradigm is diagnostic and aimed at understanding when local context fails to
override pretrained meaning, a relevant pre-deployment check for systems that must respect
document-specific definitions in domains such as law, medicine, and technical documentation.
We report aggregate behavior and causal interventions rather than prompt-attack constructions.

\bibliography{custom}

\appendix

In the appendices below, code-style names in monospace (\texttt{source\_all},
\texttt{def\_subject}, \texttt{def\_target}, \texttt{query\_word}, \texttt{game\_rule},
\texttt{prompt\_style}, etc.)
correspond to column values, group labels, or template identifiers in the released CSVs and
notebook code. The body of the paper uses the corresponding paper-level terms (\emph{full source
triplet}, \emph{definition subject}, \emph{definition target}, \emph{query word},
\emph{game-rule wrapper}, \emph{prompt wrapper}, etc.).

\section{Stimulus Construction and Prompt Templates}
\label{app:stimuli}

\paragraph{Behavioral set.}
The behavioral stimulus set contains $176$ unique lexical items, each rendered under four prompt
styles, yielding $704$ item--prompt observations per model setting. Unique items are distributed
across the four conflict families as follows: $80$ antonym, $36$ arbitrary semantic remapping,
$30$ polysemy/entity remapping, and $30$ domain-definition remapping. Targets and distractors
are chosen to be single tokens under each model's tokenizer whenever possible; when not, the
target/distractor token-count difference is included as a regression covariate
(Table~\ref{tab:regression}). The full stimulus list is released in
\texttt{behavior\_\allowbreak external\_\allowbreak validity\_\allowbreak stimuli.csv}.

\paragraph{Neutral controls.}
Neutral controls replace the query word with semantically weak words drawn from a per-item
candidate pool of $6$--$8$ words, screened to be semantically weak in context (e.g.,
\emph{thing}, \emph{object}, \emph{item}, \emph{label}, \emph{symbol} for some items;
\emph{class}, \emph{field}, \emph{point}, \emph{record}, \emph{slot}, \emph{type} for items
where those generic placeholders are themselves contextually plausible). Behavioral
experiments average $K{=}6$ controls per item; mechanistic experiments select a single
representative clean prompt closest to the item's neutral mean from $K{=}8$ candidates,
giving a deterministic clean run for activation patching. The full per-item neutral lists
are released in the stimulus CSV.

\begin{table}[H]
\centering
\footnotesize
\begin{tabular*}{\columnwidth}{@{\extracolsep{\fill}}lll@{}}
\toprule
Query & Target & Distractor \\
\midrule
\multicolumn{3}{@{}l}{\emph{Antonym remapping ($n{=}80$)}} \\
\emph{small}   & \emph{big}         & \emph{tiny}    \\
\emph{large}   & \emph{small}       & \emph{big}     \\
\emph{hot}     & \emph{cold}        & \emph{warm}    \\
\emph{cold}    & \emph{hot}         & \emph{chilly}  \\
\emph{happy}   & \emph{sad}         & \emph{glad}    \\
\emph{sad}     & \emph{happy}       & \emph{unhappy} \\
\midrule
\multicolumn{3}{@{}l}{\emph{Arbitrary semantic remapping ($n{=}36$)}} \\
\emph{apple}   & \emph{river}       & \emph{fruit}    \\
\emph{banana}  & \emph{mountain}    & \emph{fruit}    \\
\emph{tiger}   & \emph{pencil}      & \emph{animal}   \\
\emph{doctor}  & \emph{forest}      & \emph{hospital} \\
\emph{teacher} & \emph{ocean}       & \emph{school}   \\
\emph{piano}   & \emph{desert}      & \emph{music}    \\
\midrule
\multicolumn{3}{@{}l}{\emph{Polysemy / entity remapping ($n{=}30$)}} \\
\emph{apple}   & \emph{music}       & \emph{iPhone}  \\
\emph{Python}  & \emph{programming} & \emph{snake}   \\
\emph{Java}    & \emph{programming} & \emph{coffee}  \\
\emph{jaguar}  & \emph{car}         & \emph{animal}  \\
\emph{mercury} & \emph{metal}       & \emph{planet}  \\
\emph{bass}    & \emph{music}       & \emph{fish}    \\
\midrule
\multicolumn{3}{@{}l}{\emph{Domain-definition remapping ($n{=}30$)}} \\
\emph{port}    & \emph{socket}      & \emph{harbor}      \\
\emph{thread}  & \emph{process}     & \emph{sewing}      \\
\emph{cell}    & \emph{spreadsheet} & \emph{biology}     \\
\emph{token}   & \emph{symbol}      & \emph{coin}        \\
\emph{driver}  & \emph{software}    & \emph{person}      \\
\emph{kernel}  & \emph{system}      & \emph{seed}        \\
\bottomrule
\end{tabular*}
\caption{Six examples per conflict family. The conflict definition takes the form
\texttt{<Query>~means~<Target>} in the \texttt{game\_rule} wrapper, with the other three
wrappers paraphrasing it (Table~\ref{tab:prompt_templates}); the full stimulus set ($176$
items) is released in \texttt{behavior\_\allowbreak external\_validity\_stimuli.\allowbreak csv}.}
\label{tab:stim_examples}
\end{table}

\paragraph{Prompt templates.}
The four prompt wrappers are listed in Table~\ref{tab:prompt_templates}. Each wrapper
embeds the conflict definition and the query in a different framing. Behavioral experiments
instantiate all four wrappers; mechanistic experiments use the \texttt{game\_rule} wrapper from
the same template family.

\begin{table}[H]
\centering
\footnotesize
\begin{tabular*}{\columnwidth}{@{}p{\columnwidth}@{}}
\toprule
\textbf{\texttt{game\_rule}} \\
\texttt{Complete the answer with one word or phrase only.}\\
\texttt{In this game, \{subject\} means \{meaning\}.}\\
\texttt{Question: In this game, \{relation\_phrase\} \{query\} is}\\
\texttt{Answer:}\\
\midrule
\textbf{\texttt{glossary}} \\
\texttt{A glossary for this document defines "\{subject\}" as "\{meaning\}".}\\
\texttt{Using only this glossary, \{relation\_phrase\} "\{query\}" is}\\
\midrule
\textbf{\texttt{technical\_document}} \\
\texttt{In the following technical document, the term "\{subject\}" refers to "\{meaning\}".}\\
\texttt{According to the document, \{relation\_phrase\} "\{query\}" is}\\
\midrule
\textbf{\texttt{definition\_scope}} \\
\texttt{For this passage only, interpret "\{subject\}" as "\{meaning\}".}\\
\texttt{Under this definition, \{relation\_phrase\} "\{query\}" is}\\
\bottomrule
\end{tabular*}
\caption{Prompt wrappers. \{subject\} and \{meaning\} are filled in from the conflict
definition; \{query\} is the query word; \{relation\_phrase\} is ``a synonym of'' for the
antonym family and ``a word related to'' for the other three conflict families. The wrappers correspond to
the prose names \emph{game-rule}, \emph{glossary}, \emph{technical-document}, and
\emph{scoped-definition}.}
\label{tab:prompt_templates}
\end{table}

\FloatBarrier
\section{Model Set, Tokenization, and Scoring}
\label{app:models}

Table~\ref{tab:model_set} lists all model settings, with a flag indicating whether each is used
for the behavioral analysis, the mechanistic analysis, or both. The $7$B/$9$B-class models are
included for behavioral external validity. Component-level activation patching is restricted to
the tractable subset (Qwen2.5-1.5B base/IT, Gemma-2-2B base/IT, OLMo-1B), where full
residual and component patching is computationally feasible and TransformerLens
\citep{nanda2022transformerlens} loading is reliable.

\begin{table}[H]
\centering
\footnotesize
\setlength{\tabcolsep}{3pt}
\begin{tabular*}{\columnwidth}{@{\extracolsep{\fill}}llcc@{}}
\toprule
Model & Family & Beh. & Mech. \\
\midrule
Qwen2.5-1.5B           & Qwen    & \checkmark & \checkmark \\
Qwen2.5-1.5B-IT        & Qwen    & \checkmark & \checkmark \\
Qwen2.5-7B             & Qwen    & \checkmark &           \\
Qwen2.5-7B-IT          & Qwen    & \checkmark &           \\
Gemma-2-2B             & Gemma   & \checkmark & \checkmark \\
Gemma-2-2B-IT          & Gemma   & \checkmark & \checkmark \\
Gemma-2-9B             & Gemma   & \checkmark &           \\
Gemma-2-9B-IT          & Gemma   & \checkmark &           \\
OLMo-1B                & OLMo    & \checkmark & \checkmark \\
Mistral-7B             & Mistral & \checkmark &           \\
Mistral-7B-IT          & Mistral & \checkmark &           \\
\bottomrule
\end{tabular*}
\caption{Open-weight model settings. ``IT'' is the instruction-tuned variant. ``Beh.'' marks
inclusion in the behavioral sweep ($n{=}11$); ``Mech.'' marks inclusion in component-level
mechanistic analysis ($n{=}5$).}
\label{tab:model_set}
\end{table}

\paragraph{Tokenization.}
Target and distractor tokenization is identical across conflict and neutral conditions; the
query word can differ by design. The controlled regression includes target/distractor
token-count and frequency covariates (Table~\ref{tab:regression}).

\FloatBarrier
\section{Behavioral Results: Full Tables}
\label{app:behavior_tables}

\paragraph{Aggregate effect by family and prompt wrapper.}
Table~\ref{tab:agg_family} reports the aggregate sum-log-probability
effect $\effect$ at the conflict-family and prompt-wrapper levels. The aggregate effect is
positive across all four conflict families and all four prompt wrappers. Individual
model$\times$family and model$\times$wrapper cells reveal heterogeneity, especially for
antonym remapping in larger instruction-tuned models (Section~\ref{sec:behavior},
Fig.~\ref{fig:heatmaps}); the per-cell tables are released as
\texttt{behavior\_summary\_by\_\allowbreak model\_\allowbreak conflict\_\allowbreak type.csv}
and \texttt{behavior\_summary\_by\_\allowbreak model\_\allowbreak prompt\_\allowbreak style.csv}.

\begin{table}[H]
\centering
\footnotesize
\begin{tabular*}{\columnwidth}{@{\extracolsep{\fill}}lrrrr@{}}
\toprule
 & $n$ & $\effect$ & $95\%$ CI & Pos. \\
\midrule
\multicolumn{5}{@{}l}{\emph{By conflict family}} \\
Antonym       & 3520 & $1.16$ & $[1.06, 1.26]$ & $0.70$ \\
Arbitrary     & 1584 & $3.97$ & $[3.79, 4.15]$ & $0.92$ \\
Polysemy/ent. & 1320 & $2.66$ & $[2.49, 2.84]$ & $0.86$ \\
Domain-def.   & 1320 & $2.54$ & $[2.37, 2.73]$ & $0.84$ \\
\midrule
\multicolumn{5}{@{}l}{\emph{By prompt wrapper}} \\
Game-rule           & 1936 & $3.15$ & $[2.92, 3.38]$ & $0.78$ \\
Glossary            & 1936 & $2.06$ & $[1.92, 2.19]$ & $0.81$ \\
Technical-document  & 1936 & $1.75$ & $[1.64, 1.88]$ & $0.80$ \\
Scoped-definition   & 1936 & $1.94$ & $[1.82, 2.06]$ & $0.80$ \\
\bottomrule
\end{tabular*}
\caption{Aggregate Stroop interference $\effect$ by conflict family and by prompt wrapper
(sum log-probability). ``Pos.''\ is the fraction of item--prompt--model observations with
positive $\effect$.}
\label{tab:agg_family}
\end{table}

\paragraph{Per-model behavior.}
Table~\ref{tab:per_model} reports the model-level aggregate $\effect$ for the $11$ model
settings under sum log-probability scoring (Fig.~\ref{fig:forest} in the body shows the same
data graphically). All $11$ settings are positive at the model level.

\begin{table}[H]
\centering
\footnotesize
\setlength{\tabcolsep}{3pt}
\begin{tabular*}{\columnwidth}{@{\extracolsep{\fill}}lrrrr@{}}
\toprule
Model & Size & $\effect$ & $95\%$ CI & Pos. \\
\midrule
OLMo-1B           & 1.0  & $2.36$ & $[2.24, 2.48]$ & $0.95$ \\
Qwen2.5-1.5B      & 1.5  & $2.31$ & $[2.17, 2.47]$ & $0.92$ \\
Qwen2.5-1.5B-IT   & 1.5  & $2.58$ & $[2.31, 2.86]$ & $0.80$ \\
Gemma-2-2B        & 2.0  & $1.67$ & $[1.40, 1.93]$ & $0.71$ \\
Gemma-2-2B-IT     & 2.0  & $2.56$ & $[2.20, 2.92]$ & $0.75$ \\
Qwen2.5-7B        & 7.0  & $2.71$ & $[2.56, 2.86]$ & $0.93$ \\
Qwen2.5-7B-IT     & 7.0  & $1.72$ & $[1.44, 2.01]$ & $0.72$ \\
Mistral-7B        & 7.0  & $2.67$ & $[2.51, 2.84]$ & $0.91$ \\
Mistral-7B-IT     & 7.0  & $2.26$ & $[1.90, 2.63]$ & $0.66$ \\
Gemma-2-9B        & 9.0  & $2.33$ & $[2.11, 2.55]$ & $0.81$ \\
Gemma-2-9B-IT     & 9.0  & $1.31$ & $[1.00, 1.64]$ & $0.61$ \\
\bottomrule
\end{tabular*}
\caption{Model-level Stroop interference $\effect$ (sum log-probability), $n{=}704$ item--prompt
observations per model. Mean per-token scoring gives the same conclusion that all $11$ settings
are positive at the model level (released in \texttt{behavior\_summary\_by\_model.csv}).}
\label{tab:per_model}
\end{table}

\paragraph{Per-cell breakdowns.}
Tables~\ref{tab:model_x_family} and \ref{tab:model_x_style} report per-cell Stroop interference
$\effect$ for each model$\times$conflict-family and model$\times$prompt-wrapper combination,
respectively (sum log-probability). These are the numerical counterparts of the heatmaps in
Fig.~\ref{fig:heatmaps}; they make the heterogeneity discussed in
Section~\ref{sec:behavior} fully auditable.

\begin{table}[H]
\centering
\scriptsize
\setlength{\tabcolsep}{2pt}
\begin{tabular*}{\columnwidth}{@{\extracolsep{\fill}}lrrrr@{}}
\toprule
Model & Ant. & Arb. & Pol. & Dom. \\
\midrule
OLMo-1B           & $2.12$ & $3.07$ & $2.45$ & $2.06$ \\
Qwen2.5-1.5B      & $1.46$ & $4.03$ & $2.69$ & $2.15$ \\
Qwen2.5-1.5B-IT   & $1.69$ & $4.72$ & $2.41$ & $2.54$ \\
Gemma-2-2B        & $1.21$ & $2.05$ & $1.81$ & $2.28$ \\
Gemma-2-2B-IT     & $0.55$ & $6.33$ & $3.04$ & $2.87$ \\
Qwen2.5-7B        & $2.24$ & $4.09$ & $2.78$ & $2.21$ \\
Qwen2.5-7B-IT     & $0.42$ & $2.40$ & $3.39$ & $2.73$ \\
Mistral-7B        & $1.87$ & $3.94$ & $3.32$ & $2.67$ \\
Mistral-7B-IT     & $0.34$ & $5.06$ & $2.83$ & $3.46$ \\
Gemma-2-9B        & $1.46$ & $3.56$ & $2.50$ & $3.02$ \\
Gemma-2-9B-IT     & $-0.63$ & $4.38$ & $2.10$ & $2.00$ \\
\bottomrule
\end{tabular*}
\caption{Per-cell $\effect$ by model and conflict family (sum log-probability), corresponding to
Fig.~\ref{fig:heatmaps}(a). Columns: Ant.\ = antonym; Arb.\ = arbitrary semantic;
Pol.\ = polysemy/entity; Dom.\ = domain-definition. The four smallest antonym cells are all
instruction-tuned settings, and the only negative cell in the table is Gemma-2-9B-IT on antonym
remapping.}
\label{tab:model_x_family}
\end{table}

\begin{table}[H]
\centering
\scriptsize
\setlength{\tabcolsep}{2pt}
\begin{tabular*}{\columnwidth}{@{\extracolsep{\fill}}lrrrr@{}}
\toprule
Model & Game & Gloss. & Tech. & Scope \\
\midrule
OLMo-1B           & $2.41$ & $2.29$ & $2.66$ & $2.07$ \\
Qwen2.5-1.5B      & $2.97$ & $2.03$ & $2.21$ & $2.05$ \\
Qwen2.5-1.5B-IT   & $3.24$ & $2.65$ & $2.75$ & $1.67$ \\
Gemma-2-2B        & $3.49$ & $1.51$ & $0.43$ & $1.24$ \\
Gemma-2-2B-IT     & $2.69$ & $3.39$ & $1.64$ & $2.51$ \\
Qwen2.5-7B        & $3.95$ & $2.08$ & $2.48$ & $2.31$ \\
Qwen2.5-7B-IT     & $1.38$ & $2.30$ & $0.77$ & $2.44$ \\
Mistral-7B        & $4.40$ & $1.78$ & $2.65$ & $1.87$ \\
Mistral-7B-IT     & $5.00$ & $0.87$ & $0.82$ & $2.36$ \\
Gemma-2-9B        & $1.68$ & $3.01$ & $3.23$ & $1.41$ \\
Gemma-2-9B-IT     & $3.46$ & $0.72$ & $-0.32$ & $1.37$ \\
\bottomrule
\end{tabular*}
\caption{Per-cell $\effect$ by model and prompt wrapper (sum log-probability), corresponding to
Fig.~\ref{fig:heatmaps}(b). Columns: Game = game-rule; Gloss.\ = glossary;
Tech.\ = technical-document; Scope = scoped-definition.}
\label{tab:model_x_style}
\end{table}

\FloatBarrier
\section{Regression Robustness}
\label{app:regression}

Table~\ref{tab:regression} (main paper) reports the controlled OLS regression with standard
errors clustered by item. Full coefficient tables covering all conflict-family, prompt-wrapper,
and model-family fixed effects, as well as
$(\textsc{conflict\_family} \times \textsc{prompt\_style})$ and
$(\textsc{lexical\_advantage} \times \textsc{conflict\_family})$ interaction specifications, are
released in \texttt{behavior\_control\_regressions.csv}; the fixed effects and interactions are
omitted from the main table for space.

Four robustness specifications are released in
\texttt{behavior\_regression\_robustness.csv}: (i) standard errors clustered by item,
(ii) clustered by model, (iii) two-way clustered by item and model, and (iv) a linear
mixed-effects specification with item- and model-level random effects.
Table~\ref{tab:regression_robustness} summarizes these specifications for the key
\textsc{lexical\_advantage} predictor under sum log-probability scoring; the coefficient remains
positive across all four specifications (cf.\ \S\ref{sec:lexical_predictor}). Mean
log-probability scoring gives the same qualitative pattern.

\begin{table}[H]
\centering
\footnotesize
\begin{tabular*}{\columnwidth}{@{\extracolsep{\fill}}lrrr@{}}
\toprule
SE specification & Coef. & $95\%$ CI & $p$ \\
\midrule
Item-clustered          & $+0.114$ & $[+0.08, +0.15]$ & $4.5{\cdot}10^{-10}$ \\
Model-clustered         & $+0.114$ & $[+0.07, +0.16]$ & $3.7{\cdot}10^{-6}$ \\
Two-way clustered       & $+0.114$ & $[+0.06, +0.17]$ & $4.8{\cdot}10^{-5}$ \\
Mixed-effects           & $+0.114$ & $[+0.01, +0.21]$ & $0.025$ \\
\bottomrule
\end{tabular*}
\caption{Robustness of the \textsc{lexical\_advantage} coefficient under sum log-probability
scoring ($n{=}7{,}744$) across four standard-error specifications.}
\label{tab:regression_robustness}
\end{table}

\begin{table}[H]
\centering
\scriptsize
\setlength{\tabcolsep}{3pt}
\begin{tabular*}{\columnwidth}{@{\extracolsep{\fill}}lrrr@{}}
\toprule
Family & Coef. & $95\%$ CI & $p$ \\
\midrule
Antonym (baseline)        & $+0.026$ & $[-0.017, +0.070]$ & $0.235$ \\
Arbitrary semantic        & $+0.202$ & $[+0.13, +0.28]$   & $1.5{\cdot}10^{-7}$ \\
Domain-definition         & $+0.185$ & $[+0.09, +0.28]$   & $1.5{\cdot}10^{-4}$ \\
Polysemy/entity           & $+0.150$ & $[+0.08, +0.22]$   & $5.8{\cdot}10^{-5}$ \\
\bottomrule
\end{tabular*}
\caption{Family-specific lexical-prior slopes from the
$\textsc{lexical\_advantage} \times \textsc{conflict\_family}$ interaction specification
(released as \texttt{behavior\_control\_regressions.csv}, regression
\texttt{type\_interactions}). The antonym baseline slope is null; the other three families show
positive lexical-prior effects.}
\label{tab:regression_interactions}
\end{table}

\FloatBarrier
\section{Mechanistic Analysis: Implementation}
\label{app:mech_impl}

\paragraph{Clean and corrupted prompts.}
The clean (neutral remapping) and corrupted (lexical-conflict remapping) prompts share the same
target and distractor (\S\ref{sec:mech_setup}). The clean prompt is the one of the $K{=}8$
neutral candidates whose target-versus-distractor score is closest to the item's neutral mean,
giving a deterministic clean run for activation patching.

\paragraph{Mechanistic item selection.}
An antonym item enters the mechanistic set for a given model under three criteria: its query
word, target, and distractor are each a single token under that model's tokenizer; its clean and
corrupted prompts have equal token length, so that positions align without padding; and its
item-level interference is positive for that model. Up to $32$ eligible items are used per model.
The first two criteria are properties of the tokenizer and the template. The third conditions the
mechanistic sample on the behavioral outcome, and the metric requires it: recovery normalizes by
the clean$-$corrupt gap, and for an item with no interference to repair that gap is near zero or
negative, which leaves $R$ uninterpretable. The third criterion also makes the sample size
model-dependent, giving $32$ items for Qwen2.5-1.5B base and IT, Gemma-2-2B, and OLMo-1B and $14$
for Gemma-2-2B-IT, which has the weakest antonym effect of the five mechanism models
(Table~\ref{tab:model_x_family}). The selection relaxes the interference criterion if fewer than
ten items would survive it; no model reached that threshold.

\paragraph{Residual-stream sites.}
We refer to three sites within each transformer block: block input (CSV column
\texttt{hook\_resid\_pre}), mid-block, i.e., post-attention and pre-MLP
(\texttt{hook\_resid\_mid}), and block output (\texttt{hook\_resid\_post}). The main paper
uses the readable names; CSV column names follow the TransformerLens convention.

\paragraph{Source positions.}
Source positions are named \texttt{def\_subject}, \texttt{def\_target}, and
\texttt{query\_word} in the released CSVs; the full triplet intervention is named
\sourceall. The \emph{final position} is the final prompt position at which the
target-minus-distractor logit difference is read. Although the triplet combines three source
positions, the binding signal at this resolution localizes primarily to \texttt{def\_target}
(\S\ref{sec:mechanism}).

\paragraph{Recovery.}
The patching score is the final-position target-minus-distractor logit difference
$M = \ell_t - \ell_d$. Recovery is
$R = (M_{\text{patched}} - M_{\text{corrupt}}) / (M_{\text{clean}} - M_{\text{corrupt}})$,
the standard normalized metric of \citet{zhang2024towards}: $R{=}1$ is full clean--corrupt
recovery, $R{=}0$ is no recovery, and $R$ can exceed $1$ or fall below $0$ when the patch
overshoots or actively harms the answer state.

\paragraph{Split validation.}
For each model we run $200$ repeated splits. Each split chooses a discovery item set, selects
the best (hook, layer, source-group) patch on the discovery items, and evaluates that patch on the
held-out items, recording the held-out $\Delta R = R_{\text{triplet}} - R_{\text{best pair}}$.
Held-out splits in base models have positive $\Delta R$ in $0.97$--$1.00$ of splits.
Instruction-tuned models have a positive split mean with lower positive fractions and
split-level intervals that cross zero (Table~\ref{tab:split_summary}).

\paragraph{Item-mismatched source controls.}
For each model we run $50$ control perturbations. Each perturbation patches clean source
activations from a randomly chosen \emph{different} item, leaving the corrupted item's clean
target-distractor pair fixed. These controls test whether recovery requires item-matched source
information; they do not recover behavior and often actively harm it
(Table~\ref{tab:random_summary}).

\FloatBarrier
\section{Additional Mechanistic Results}
\label{app:mech_extra}

\paragraph{Source groups at the triplet-selected site and layer.}
Table~\ref{tab:source_group_recovery} reports the recovery of each source group evaluated at
the \emph{same residual-stream site and layer} that the full triplet selects in
Table~\ref{tab:mechsummary} for that model. This is the comparison point used by the main
source-triplet analysis: at each model's selected site and layer, the full triplet exceeds
every singleton and every pair. The released CSV also contains all per-layer values for the
three residual-stream sites: block input, mid-block, and block output.

\begin{table}[H]
\centering
\scriptsize
\setlength{\tabcolsep}{2pt}
\begin{tabular*}{\columnwidth}{@{\extracolsep{\fill}}lrrrrr@{}}
\toprule
Group & Q1.5 & Q1.5-IT & G2 & G2-IT & OLMo \\
(site, $L$) & input,$\,3$ & input,$\,15$ & mid-block,$\,12$ & input,$\,3$ & input,$\,2$ \\
\midrule
def\_s                            & $-2.46$ & $-0.09$ & $+0.07$ & $-3.33$ & $-0.02$ \\
def\_t                            & $+0.03$ & $-1.99$ & $+0.22$ & $+0.34$ & $+0.06$ \\
query                             & $-2.35$ & $+0.35$ & $+0.30$ & $-1.20$ & $-0.58$ \\
def\_s\,+\,def\_t                 & $-2.31$ & $-1.92$ & $+0.32$ & $-3.75$ & $+0.09$ \\
def\_s\,+\,query                  & $+0.89$ & $+0.59$ & $+0.69$ & $+0.68$ & $+0.87$ \\
def\_t\,+\,query                  & $-2.20$ & $-0.09$ & $+0.59$ & $-1.02$ & $-0.50$ \\
\textbf{Triplet}                  & $\mathbf{0.99}$ & $\mathbf{0.92}$ & $\mathbf{1.05}$ & $\mathbf{1.04}$ & $\mathbf{1.06}$ \\
\bottomrule
\end{tabular*}
\caption{Source-group recoveries at each model's triplet-selected residual-stream site and
layer from Table~\ref{tab:mechsummary}. At the selected site and layer, the full triplet
exceeds every singleton and every pair in all five models. Column headings: Q1.5 =
Qwen2.5-1.5B; Q1.5-IT = Qwen2.5-1.5B-IT; G2 = Gemma-2-2B; G2-IT = Gemma-2-2B-IT; OLMo =
OLMo-1B. Row headings: def\_s = definition subject; def\_t = definition target; query =
query word.}
\label{tab:source_group_recovery}
\end{table}

\paragraph{Own-best-layer protocol.}
Evaluating each source group at its own-best (site, layer) rather than the triplet-selected
position lets the definition subject\,+\,query word pair reach high recovery at early layers,
where the patch approaches an input replacement. Those two positions are the only ones at which
the clean and corrupted token sequences differ (\S\ref{sec:mech_setup}), so copying both at
near-embedding depth restores the entire input difference between the runs, and the
definition-target position has not yet accumulated the context that makes it informative. We therefore use the matched-site/layer comparison in
Table~\ref{tab:source_group_recovery} as the main comparison point and the held-out split
validation (Table~\ref{tab:split_summary}) as the check that does not inherit that choice. All
per-group per-layer
recoveries (and all three residual-stream sites: block input, mid-block, and block output)
are in \texttt{mechanism\_source\_residual\_\allowbreak patches.csv}.

\paragraph{Definition-target swap control.}
Donor definition-target activations are drawn from the same matched pool used for the
matched-triplet patch. We run $10$ donor swaps per item at each model's triplet-selected
site and layer and report, in Table~\ref{tab:swap_summary}, the matched recovery, the mean
swap recovery, and the fraction of items whose mean swap recovery falls below the matched
value (\emph{every} item in every model). The near-identity between swap and item-mismatched
outcomes at the margin level (\S\ref{sec:mechanism}, Table~\ref{tab:logit_decomp_all})
further indicates that the binding signal at this resolution is carried almost entirely by
the \texttt{def\_target} position.

\begin{table}[H]
\centering
\scriptsize
\setlength{\tabcolsep}{2pt}
\begin{tabular*}{\columnwidth}{@{\extracolsep{\fill}}lcrrrr@{}}
\toprule
Model & (site, $L$) & $n$ & Matched & Swap & Frac.\,below \\
\midrule
Qwen2.5-1.5B    & input,$\,3$     & $32$ & $+0.99$ & $-4.26$ & $1.00$ \\
Qwen2.5-1.5B-IT & input,$\,15$    & $32$ & $+0.92$ & $-0.62$ & $1.00$ \\
Gemma-2-2B      & mid-block,$\,12$ & $32$ & $+1.05$ & $-0.54$ & $1.00$ \\
Gemma-2-2B-IT   & input,$\,3$     & $14$ & $+1.04$ & $-2.43$ & $1.00$ \\
OLMo-1B         & input,$\,2$     & $32$ & $+1.06$ & $-1.14$ & $1.00$ \\
\bottomrule
\end{tabular*}
\caption{Definition-target swap control
(\texttt{mechanism\_definition\_target\_swap\_\allowbreak summary.csv}). The definition subject
and query word activations are kept matched to the item; only the definition-target activation
is replaced by a donor item's definition-target activation. Matched is the matched
triplet recovery; Swap is the mean swap recovery across $10$ donor swaps per item;
Frac.\,below is the fraction of items whose mean swap recovery is below the matched
triplet recovery. Recovery drops sharply in every model, supporting the binding
interpretation.}
\label{tab:swap_summary}
\end{table}

\paragraph{Reader zero-ablation mediation.}
A non-degenerate component scan at the final position (attention heads and MLPs whose
individual-component patch effect is reproducible across split seeds and not subsumed by
trivial null perturbations) identifies, in each mechanism model, a small set of \emph{reader}
components that mediate the triplet patch. Table~\ref{tab:mediation} lists the top reader
components by \emph{mediation drop}: the reduction in triplet-patch recovery when the listed
component is zero-ablated during the matched triplet patch. The top reader in every model
contributes a recovery drop of at least $0.19$, and individual readers account for drops of
up to $0.62$. Per-model writer scans at source positions are released in
\texttt{mechanism\_component\_scan\_\allowbreak nondegenerate.csv}; writer effects are smaller
and more distributed across heads/MLPs than the reader effects reported here.
Not every non-degenerate component supports recovery: $8$ components (across the five models),
primarily late-block MLPs, yield a negative drop of magnitude $\geq 0.3$, meaning that
zero-ablating them \emph{increases} patched recovery. The pattern is reproducible across
split seeds and is consistent with a suppression-style role for these components, in which
removing them releases the patched target--distractor margin upward (cf.\ copy suppression,
\citealp{mcdougall2024copy}); the full list is in the component-mediation CSV.

\begin{table}[H]
\centering
\footnotesize
\begin{tabular*}{\columnwidth}{@{\extracolsep{\fill}}llrrr@{}}
\toprule
Model & Comp. & $R_{\text{trp}}$ & $R_{\text{trp+0}}$ & Drop \\
\midrule
Qwen2.5-1.5B    & L20H5  & $0.99$ & $0.37$ & $0.62$ \\
Qwen2.5-1.5B    & L26H7  & $0.99$ & $0.44$ & $0.56$ \\
Qwen2.5-1.5B-IT & L27MLP & $0.92$ & $0.56$ & $0.36$ \\
Qwen2.5-1.5B-IT & L20H5  & $0.92$ & $0.63$ & $0.29$ \\
Gemma-2-2B      & L12H2  & $1.05$ & $0.79$ & $0.26$ \\
Gemma-2-2B      & L22H7  & $1.05$ & $0.80$ & $0.24$ \\
Gemma-2-2B-IT   & L22H1  & $1.04$ & $0.58$ & $0.47$ \\
Gemma-2-2B-IT   & L16H4  & $1.04$ & $0.72$ & $0.33$ \\
OLMo-1B         & L8MLP  & $1.06$ & $0.78$ & $0.28$ \\
OLMo-1B         & L15H12 & $1.06$ & $0.87$ & $0.19$ \\
\bottomrule
\end{tabular*}
\caption{Top reader zero-ablation drops per model
(\texttt{mechanism\_component\_mediation\_\allowbreak zero\_ablation.csv}). $R_{\text{trp}}$ is
the matched triplet recovery; $R_{\text{trp+0}}$ is the triplet recovery with the listed
component zero-ablated; \emph{Drop} = $R_{\text{trp}} - R_{\text{trp+0}}$.}
\label{tab:mediation}
\end{table}

\paragraph{Split-validation summary.}
Table~\ref{tab:split_summary} reports the $200$-fold split-validation results per mechanism
model: held-out mean $\Delta R = R_{\text{triplet}} - R_{\text{pair}}$, $95\%$ split-level
confidence interval, and fraction of splits with positive $\Delta R$ (cf.\ \S\ref{sec:mechanism}).

\begin{table}[H]
\centering
\scriptsize
\setlength{\tabcolsep}{2pt}
\begin{tabular*}{\columnwidth}{@{\extracolsep{\fill}}lrrcr@{}}
\toprule
Model & Splits & Mean $\Delta R$ & $95\%$ CI & Pos. \\
\midrule
Qwen2.5-1.5B    & $200$ & $+0.11$ & $[+0.00,\,+0.34]$ & $0.975$ \\
Qwen2.5-1.5B-IT & $200$ & $+0.31$ & $[-0.01,\,+0.56]$ & $0.950$ \\
Gemma-2-2B      & $200$ & $+0.39$ & $[+0.03,\,+0.66]$ & $0.975$ \\
Gemma-2-2B-IT   & $200$ & $+0.19$ & $[-0.11,\,+0.58]$ & $0.825$ \\
OLMo-1B         & $200$ & $+0.18$ & $[+0.08,\,+0.28]$ & $1.000$ \\
\bottomrule
\end{tabular*}
\caption{Held-out split-validation summary per mechanism model
(\texttt{mechanism\_split\_validation.csv}, summary in
\texttt{mechanism\_split\_validation\_\allowbreak summary.csv}). \textbf{Pos.}\ is the fraction
of splits with positive $\Delta R$.}
\label{tab:split_summary}
\end{table}

\paragraph{Item-mismatched source controls summary.}
Table~\ref{tab:random_summary} summarizes the $50$ item-mismatched control perturbations per
mechanism model. The matched triplet recovery is near or above $1$ for every model, while
the mean control recovery is strongly negative; the highest single control value
(\textbf{Ctrl.\,max}) does not approach the matched value in any model.

\begin{table}[H]
\centering
\scriptsize
\setlength{\tabcolsep}{2pt}
\begin{tabular*}{\columnwidth}{@{\extracolsep{\fill}}lrrrr@{}}
\toprule
Model & Matched & Ctrl.\,mean & Ctrl.\,min & Ctrl.\,max \\
\midrule
Qwen2.5-1.5B    & $0.99$ & $-4.16$ & $-4.90$ & $-3.35$ \\
Qwen2.5-1.5B-IT & $0.92$ & $-0.66$ & $-0.88$ & $-0.40$ \\
Gemma-2-2B      & $1.05$ & $-0.58$ & $-0.80$ & $-0.30$ \\
Gemma-2-2B-IT   & $1.04$ & $-2.27$ & $-3.10$ & $-1.33$ \\
OLMo-1B         & $1.06$ & $-1.16$ & $-1.38$ & $-0.87$ \\
\bottomrule
\end{tabular*}
\caption{Item-mismatched source-control summary per mechanism model
(\texttt{mechanism\_random\_controls.csv}; $50$ permutations per model). The matched triplet
recovery is far above the worst-case control max in every model.}
\label{tab:random_summary}
\end{table}

\FloatBarrier
\section{Reproducibility and Released Artifacts}
\label{app:reproducibility}

\paragraph{Code, data, and repository.}
The complete implementation, the end-to-end reproduction notebook, and the released CSV
outputs, figures, Markdown summaries, and package documentation are available at
\url{https://github.com/henryhyw/priors-persist-through-suppression}. The notebook reproduces all numerical results reported in the paper; the
repository's \texttt{README} indexes each CSV against the figure or table it supports.

\paragraph{Artifacts, licenses, and intended use.}
The released code and documentation are distributed for research reproduction and diagnostic
analysis. The code is MIT-licensed. Pretrained model checkpoints, TransformerLens, and Hugging
Face tooling retain their original licenses, access conditions, and model-card terms; gated
checkpoints require the corresponding upstream access approval. The created artifacts are not
intended for deployment or downstream decision-making.

\paragraph{Data, human subjects, and risk profile.}
The released data consist of synthetic prompt templates, lexical-item lists, generated CSV
outputs, figures, and Markdown summaries. No human subjects, crowdworkers, private user data,
or sensitive personal data were collected. Because the paradigm is diagnostic, the main risk is
over-interpreting lexical-override interference as a general prompt-attack or safety-control
method; we report aggregate behavior and causal interventions rather than operational attack
instructions.

\paragraph{Hardware and precision.}
Reproduction runs on a single A100-class GPU and involves forward evaluation plus
activation caching and patching, with no model training or fine-tuning. Behavioral evaluation
uses \texttt{bfloat16} where available; mechanistic experiments use \texttt{float32} for stable
activation caching and consistent residual-stream arithmetic across model families.

\paragraph{Tooling.}
Mechanistic interventions use TransformerLens \citep{nanda2022transformerlens}. Gated model
checkpoints (e.g., the Gemma family) require a Hugging Face access token; the notebook reads
it from the \texttt{HF\_TOKEN} environment variable.

\paragraph{Tokenizer handling.}
Some open-weight tokenizers (notably the Qwen family) require explicit handling because the
default behavior assumes a beginning-of-sequence token that is not present in the vocabulary.
We disable automatic BOS insertion when the tokenizer has no BOS token and use explicit
\texttt{add\_special\_tokens=False} scoring, ensuring consistent prompt likelihood computation
across model families.

\paragraph{Determinism.}
Behavioral evaluation is fully deterministic given fixed model weights and tokenizer handling.
Mechanistic split validation and item-mismatched source controls use a fixed random seed
recorded in the notebook; the released CSVs are the exact outputs of that seeded run.

\end{document}